\documentclass[10pt,twocolumn,letterpaper]{article}

\usepackage{cvpr}
\usepackage{times}
\usepackage{epsfig}
\usepackage{graphicx}
\usepackage{amsmath}
\usepackage{amssymb}

\usepackage[pagebackref=false,breaklinks=true,letterpaper=true,colorlinks,bookmarks=false]{hyperref}

 \cvprfinalcopy 

\def\cvprPaperID{367} 

\ifcvprfinal\fi

\usepackage[utf8]{inputenc} 
\usepackage[T1]{fontenc}    
\usepackage{url}            
\usepackage{booktabs}       
\usepackage{amsfonts}       
\usepackage{nicefrac}       
\usepackage{microtype}      

\usepackage[nocompress]{cite}
\usepackage{amsthm}
\usepackage{stmaryrd}
\usepackage{multirow}
\usepackage{tabularx}
\usepackage{arydshln}
\usepackage{makecell}
\usepackage{lipsum}
\usepackage[lined,ruled,commentsnumbered]{algorithm2e}
\usepackage{tablefootnote}
\usepackage{threeparttable}
\usepackage{caption}
\usepackage{subcaption}
\usepackage{sidecap}
\usepackage{color}
\usepackage{stfloats}

\newcommand{\ndv}{n_{d,v}}
\newcommand{\Ndv}{N_{d,v}}
\newcommand{\cdv}{c_{d,v}}
\newcommand{\Cdv}{C_{d,v}}
\newcommand{\Oh}{O}

\newcommand{\define}{\overset{\Delta}{=}}
\newcommand{\Rd}{R^{(d)}}
\newcommand{\apt}{\mathrm{AP}_\mathrm{T}}
\newcommand{\dcgt}{\mathrm{DCG}_\mathrm{T}}
\newcommand{\apr}{\mathrm{AP}_\mathrm{r}}
\newcommand{\dcgr}{\mathrm{DCG}_\mathrm{r}}
\newcommand{\ndcgt}{\mathrm{NDCG}_\mathrm{T}}
\newcommand{\aff}{\mathcal{A}}
\newcommand{\affset}{\mathcal{V}}
\newcommand{\nbits}{b}
\newcommand{\cifar}{CIFAR-10\ }

\newcommand{\labelme}{LabelMe\ }
\newcommand{\sgn}{\mathop{\mathrm{sgn}}}

\newtheorem{prop}{Proposition}

\title{Hashing as Tie-Aware Learning to Rank}

\author{
\begin{tabular}[t]{cccc} 
Kun He ~ &  Fatih Cakir ~ &  Sarah Adel Bargal ~ &  Stan Sclaroff \\
\multicolumn{4}{c}{Department of Computer Science, Boston University}\\
\multicolumn{4}{c}{\texttt{\normalsize \{hekun,\,fcakir,\,sbargal,\,sclaroff\}@cs.bu.edu}}
\end{tabular}
}

\begin{document}
\maketitle

\begin{abstract}
Hashing, or learning binary embeddings of data, is frequently used in nearest neighbor retrieval.
In this paper, we develop learning to rank formulations for hashing, aimed at directly optimizing ranking-based evaluation metrics such as Average Precision (AP) and Normalized Discounted Cumulative Gain (NDCG).
We first observe that the integer-valued Hamming distance often leads to tied rankings, and propose to use tie-aware versions of AP and NDCG to evaluate hashing for retrieval.
Then, to optimize tie-aware ranking metrics, we derive their continuous relaxations, and perform gradient-based optimization with deep neural networks. 
Our results establish the new state-of-the-art for image retrieval by Hamming ranking in common benchmarks.
\end{abstract}

\section{Introduction}

\begin{figure}[t]
\centering
\includegraphics[width=.9\linewidth]{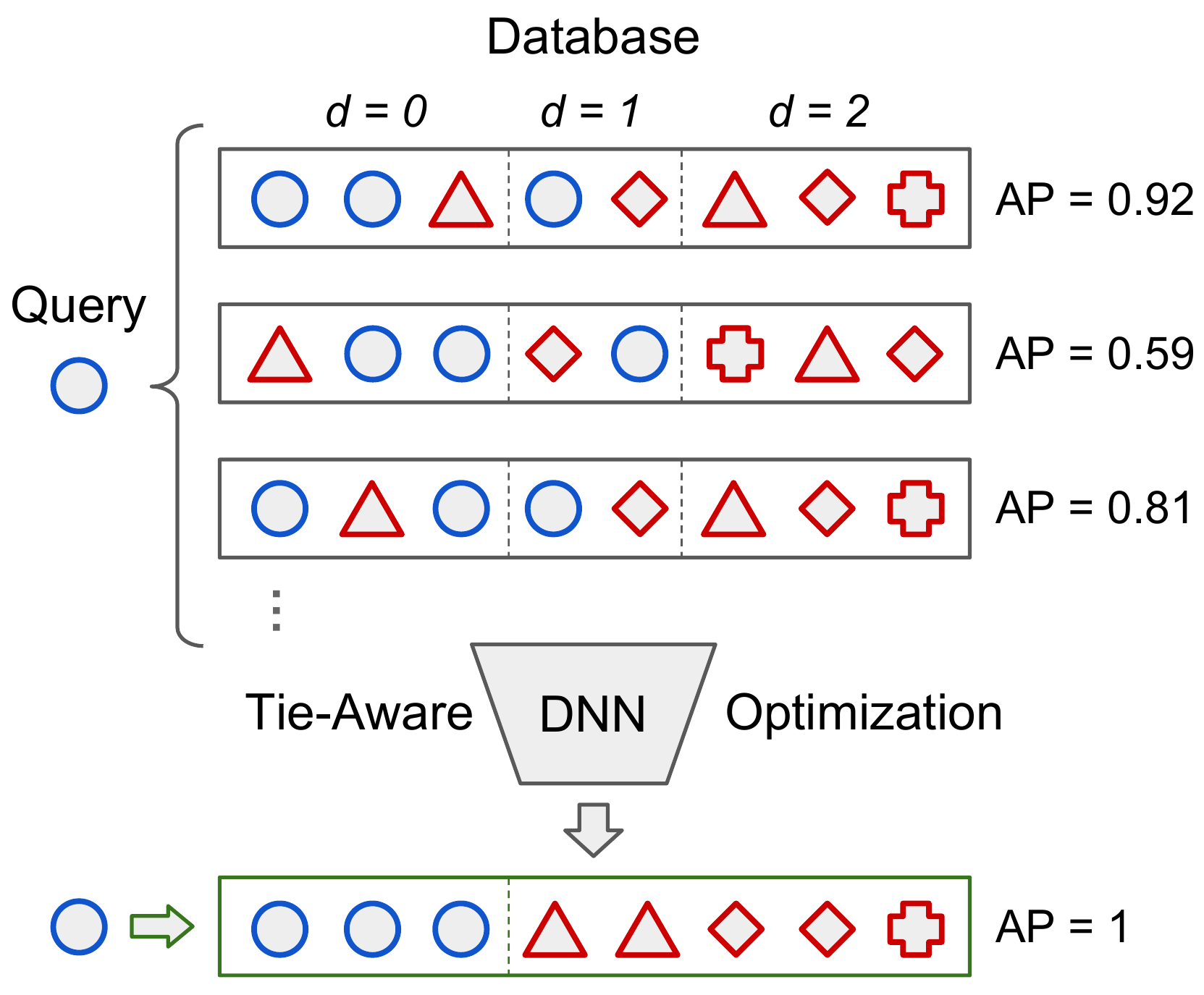}
\vspace{-.2em}
\caption{
When applying hashing for nearest neighbor retrieval, the integer-valued Hamming distance produces \emph{ties} (items that share the same distance).
If left uncontrolled, different tie-breaking strategies could give drastically different values of the evaluation metric, \eg AP.
We address this issue by using \emph{tie-aware} ranking metrics that implicitly average over all the permutations in closed form.
We further use tie-aware ranking metrics as optimization objectives in deep hashing networks, leading to state-of-the-art results.
}
\vspace{-.5em}
\label{fig:fig1}
\end{figure}

In this paper, we consider the problem of hashing, which is concerned with learning binary embeddings of data in order to enable fast approximate nearest neighbor retrieval.
We take a task-driven approach, and seek to optimize learning objectives that closely match test-time performance measures. 
Nearest neighbor retrieval performance is frequently measured using ranking-based evaluation metrics, such as Average Precision (AP) and Normalized Discounted Cumulative Gain (NDCG) \cite{introIR}, 
but the optimization of such metrics has been deemed difficult in the hashing literature \cite{affinity}.
We propose a novel learning to rank formulation to tackle these difficult optimization problems, 
and our main contribution is a gradient-based method that directly optimizes ranking metrics for hashing.
Coupled with deep neural networks, this method achieves state-of-the-art results.

Our formulation is inspired by a simple observation.
When performing retrieval with binary vector encodings and the integer-valued Hamming distance, the resulting ranking usually contains \emph{ties}, and different tie-breaking strategies can lead to different results (Fig.~\ref{fig:fig1}).
In fact, ties are a common problem in ranking, and much attention has been paid to it, including in Kendall's classical work on rank correlation \cite{Kendall1948}, and in the modern information retrieval literature \cite{ties,Cabanac2010}. 
Unfortunately, the learning to hash literature largely lacks tie-awareness, and current evaluation protocols rarely take tie-breaking into account.
Thus, we advocate using \emph{tie-aware} ranking evaluation metrics, which implicitly average over all permutations of tied items, and permit efficient closed-form evaluation.

Our natural next step is to learn hash functions by optimizing tie-aware ranking metrics.
This can be seen as an instance of learning to rank with listwise loss functions, which is advantageous compared to many other ranking-inspired hashing formulations. 
To solve the associated discrete and NP-hard optimization problems, we relax the problems into their continuous counterparts where closed-form gradients are available, and then perform gradient-based optimization with deep neural networks.
We specifically study the optimization of AP and NDCG, two ranking metrics that are widely used in evaluating nearest neighbor retrieval performance. Our results establish the new state-of-the-art for these metrics in common image retrieval benchmarks.

\section{Related Work}
\label{sec:relatedwork}

Hashing is a widely used approach for practical nearest neighbor retrieval \cite{hash_survey}, thanks to the efficiency of evaluating Hamming distances using bitwise operations, as well as the low memory and storage footprint.
It has been  theoretically demonstrated \cite{hash_optimal} that data-dependent hashing methods outperform data-independent ones such as Locality Sensitive Hashing \cite{LSH}. 
We tackle the supervised hashing problem, also known as affinity-based hashing \cite{affinity,BRE,SHK}, where supervision is given in the form of pairwise affinities.
Regarding optimization, the discrete nature of hashing usually results in NP-hard problems. 
Our solution uses continuous relaxations,  which is in line with relaxation-based methods, \eg \cite{BRE,SHK,mihash}, but differs from alternating methods that preserve the discrete constraints \cite{affinity,LinIJCV2016,HDML} and two-step methods \cite{FastHash,Zhuang2016CVPR,SHECC}.

Supervised hashing can be cast as a distance metric learning problem \cite{HDML}, which itself can be formulated as learning to rank \cite{MLR,Lim2014}.
Optimizing ranking metrics such as AP and NDCG  has received much attention in the learning to rank literature.
For instance, surrogates of AP  and NDCG can be optimized in the structural SVM framework \cite{APSVM,NDCGSVM}, and bound optimization algorithms exist for NDCG \cite{rank_ndcg}.
Alternatively, there are gradient-based methods based on smoothing or approximating these metrics \cite{smoothNDCG,Chapelle2010,softrank,lambdarank}. 
Recently, \cite{fewshotAP} tackles few-shot classification by optimizing AP using the direct loss minimization framework \cite{dlm_nn}.
These methods did not consider applications in hashing.

In the learning to hash literature, different strategies have been proposed to handle the difficulties in optimizing listwise ranking metrics.
For example, \cite{Wang2013listwise} decomposes listwise supervision into local triplets, \cite{LinIJCV2016,Yu2014SIGIR} use structural SVMs to optimize surrogate losses, \cite{toprank} maximizes precision at the top, and \cite{Wang2015IJCAI,Zhao2015CVPR} optimize NDCG surrogates.
In other recent methods using deep neural networks, the learning objectives are not designed to match ranking evaluation metrics, \eg \cite{DPSH,DTSH,mihash,Zhuang2016CVPR}.
In contrast, we directly optimize listwise ranking metrics using deep neural networks.

Key to our formulation is the observation that the integer-valued Hamming distance results in rankings with ties. However, this fact is not widely taken into consideration in previous work.
Ties can be sidestepped by using weighted Hamming distance \cite{WHD,LinIJCV2016}, but at the cost of reduced efficiency.
Fortunately, tie-aware versions of common ranking metrics have been found in the information retrieval literature \cite{ties}. 
Inspired by such results, we propose to optimize tie-aware ranking metrics on Hamming distances.
Our  gradient-based optimization uses a recent differentiable histogram binning technique \cite{mihash,cakir2018hashing,histloss}.

\section{Hashing as Tie-Aware Ranking}
\label{sec:tieaware}

\subsection{Preliminaries}
\label{sec:prelim}

\textbf{Learning to hash.} In learning to hash, we wish to learn a hash mapping $\Phi:\mathcal{X}\to \mathcal{H}^\nbits$, where $\mathcal{X}$ is the feature space, and $\mathcal{H}^\nbits=\{-1,1\}^\nbits$ is the $\nbits$-dimensional Hamming space.
A hash mapping $\Phi$ induces the Hamming distance $d_\Phi:\mathcal{X}\times\mathcal{X}\to\{0,1,\ldots,\nbits\}$ as\footnote{Although the usual implementation is by counting bit differences, this equivalent formulation has the advantage of being differentiable.}
\begin{align}
d_\Phi(x,x')=\frac{1}{2}\left(b-\Phi(x)^\top\Phi(x')\right).
\label{eq:hamming_dist}
\end{align}

We consider a supervised learning setting, or \emph{supervised hashing}, where supervision is specified using {pairwise affinities}.
Formally, we assume access to an affinity oracle $\aff$, whose value indicates  a notion of similarity: 
two examples $x_i,x_j\in\mathcal{X}$ are called similar if $\aff(x_i,x_j)>0$, and dissimilar when $\aff(x_i,x_j)=0$.
In this paper, we restrict $\aff$ to take values from a finite set $\affset$, which covers two important special cases. 
First,  $\affset=\{0,1\}$, or \emph{binary affinities}, are extensively studied in the current literature.
Binary affinities can be derived from agreement of class labels, or by thresholding the original Euclidean distance in $\mathcal{X}$.\footnote{The latter is sometimes referred to as ``unsupervised hashing'' in the literature due to the absence of class labels.}
The second case is \emph{multi-level affinities}, where $\affset$ consists of non-negative integers. This more fine-grained model of similarity is frequently considered in information retrieval tasks, including in web search engines.

Throughout this paper we assume the setup where a {query} $x_q\in\mathcal{X}$ is retrieved against some {database} $S\subseteq\mathcal{X}$. Retrieval is performed by ranking the instances in $S$ by increasing distance to $x_q$, 
using $d_\Phi$ as the distance metric.
This is termed ``retrieval by Hamming ranking'' in the hashing literature.
The ranking can be represented by an index vector $R$, whose elements form a permutation of $\{1,\ldots,|S|\}$. 
Below, let $R_i$ be the $i$-th element in $R$, and $\aff_q(i)=\aff(x_q,x_i)$.
Unless otherwise noted, we implicitly assume dependency on $x_q,S$, and $\Phi$ in our notation.

\textbf{Ranking-based evaluation.}
Ranking-based metrics usually measure some form of agreement between the ranking and ground truth affinities, capturing the intuition that retrievals with high affinity to the query should be ranked high.
First, in the case of binary affinity, we define $N^+=\left|\{x_i\in S|\aff_q(i)=1\}\right|$.
Average Precision (AP) averages the precision at cutoff $k$ over all cutoffs:
\begin{align}
\text{AP}(R)
= \frac{1}{N^+}\sum_{k=1}^{|S|} \aff_q(R_k)\left[\frac{1}{k}\sum_{j=1}^k \aff_q(R_j)\right].
\label{eq:ap}
\end{align}
Next, for integer-valued affinities, Discounted Cumulative Gain (DCG) is defined as
\begin{align}
\text{DCG}(R) & = \sum_{k=1}^{|S|}G(\aff_q(R_k))D(k), \\
\text{where}~~ G(a) & =2^a-1, ~ D(k)=\frac{1}{\log_2(k+1)}.
\label{eq:ndcg}
\end{align}
$G$ and $D$ are called gain and (logarithmic) discount, respectively. 
Normalized DCG (NDCG) divides DCG by its maximum possible value, ensuring a range of $[0,1]$:
\begin{align}
\text{NDCG}(R) = \frac{\text{DCG}(R)}{\max_{R'} \text{DCG}(R')}.
\end{align}

\begin{table*}[bp]
\hrule
\begin{prop}
\label{prop:tie}
Both $\apt$ and $\dcgt$ decompose additively over the ties. 
For $\affset=\{0,1\}$, let $n_d^+\overset{\Delta}{=} n_{d,1},N_d^+\overset{\Delta}{=} N_{d,1}$, and $N^+=\sum_{d}n_d^+$,
the contribution of each tie $R^{(d)}$ to $\apt$ is computed as
\begin{align}
\apt(R^{(d)})
= \frac{n_d^+}{n_dN^+}\sum_{t=N_{d-1}+1}^{N_d}\frac{N_{d-1}^++(t-N_{d-1}-1)\frac{n_d^+-1}{n_d-1}+1}{t}. 
\label{eq:apt}
\end{align}
For $\dcgt$, the contribution of $R^{(d)}$ is
\begin{align}
\dcgt(R^{(d)}) 
 = \sum_{i\in R^{(d)}}\frac{ G(\aff_q(i))}{n_d}\!\!\sum_{t=N_{d-1}+1}^{N_d}\!\!D(t) 
 = \sum_{v\in\affset}\frac{G(v)\ndv}{n_d}\!\!\sum_{t=N_{d-1}+1}^{N_d}\!\!D(t).
\label{eq:dcgt}
\end{align}
\begin{proof}
See appendix.
\end{proof}
\end{prop}
\hrule
\end{table*}

\subsection{Tie-Awareness in Hashing}
\label{sec:ties}

When evaluating information retrieval systems, special attention is required when there exist {ties} in the distances \cite{ties,Cabanac2010}.  
In this case, the ranking $R$ is not unique as the tied items can be ordered arbitrarily, and the tie-breaking strategy may have a sizable impact on the result.
We have given an example in Fig.~\ref{fig:fig1}.
Surprisingly, we found that current ranking-based hashing evaluation protocols usually do not take tie-breaking into account, which could result in ambiguous comparisons or even unfair exploitation.
Perhaps more importantly, ties render the formulation of direct optimization unclear: what tie-breaking strategy should we assume when using AP or NDCG as optimization objectives?
Thus, we believe that it is important to seek \emph{tie-aware} evaluation metrics for hashing.

Rather than picking a fixed tie-breaking strategy or relying on randomization, the tie-aware solution that we propose is to average the value of the ranking metric over all possible permutations of tied items.
This solution is appealing in several ways: it is deterministic, it is unambiguous and cannot be exploited, and it reduces to the ordinary version when there are no ties. 
However, there is one caveat: generating all permutations for $n$ tied items requires $O(n!)$ time, which is super-exponential and prohibitive.
Fortunately, \cite{ties} observes that the average can be computed implicitly for commonly used ranking metrics, and gives their tie-aware versions in closed form.
Based on this result, we further describe how to efficiently compute tie-aware ranking metrics by exploiting the structure of the Hamming distance.

We focus on AP and NDCG, and denote the tie-aware versions of AP and (N)DCG as $\apt$ and ($\mathrm{N}$)$\dcgt$, respectively.
First, we define some notation.
With integer-valued Hamming distances, we redefine the ranking $R$ to be a collection of $(b+1)$ ``ties'', \ie $R=\{R^{(0)},\ldots,R^{(\nbits)}\}$, where $R^{(d)}=\{i|d_\Phi(x_q,x_{i})=d\}$ is the set of retrievals having Hamming distance $d$ to the query. 
We define a set of discrete histograms conditioned on affinity values,
$(n_{0,v},\ldots,n_{\nbits,v})$, where $\ndv=|R^{(d)}\cap \{i|\aff_q(i)=v\}|, \forall v\in\affset$, and their cumulative sums $(N_{0,v},\ldots,N_{\nbits,v})$ where $\Ndv=\sum_{j\leq d}n_{j,v}$.
We also define the total histograms as $n_d=\sum_{v\in\affset}\ndv$ with cumulative sum $N_d=\sum_{j\leq d}n_j$.

Next, Proposition~\ref{prop:tie} gives the closed forms of $\apt$ and $\dcgt$.
We give proof in the appendix.

\textbf{Time complexity Analysis.} 
Let $|S|=N$.
Given the Hamming distances $\{d_\Phi(x_q,x)|x\in S\}$, the first step is to generate the ranking $R$, or populate the ties $\{R^{(d)}\}$.
This step is essentially the \emph{counting sort} for integers, which has $\Oh(\nbits N)$ time complexity.
Computing either $\apt$ or $\dcgt$ then takes $\Oh(\sum_dn_d)=\Oh(N)$ time, which makes the total time complexity $\Oh(\nbits N)$.
In our formulation, the number of bits $\nbits$ is a constant, and therefore the complexity is linear in $N$.
In contrast, for real-valued distances, sorting generally takes $\Oh(N\log N)$ time and is the dominating factor.

For the normalized $\ndcgt$, the normalizing factor is unaffected by ties, but computing it still requires sorting the gain values in descending order. 
Under the assumption that the set of affinity values $\affset$ consists of non-negative integers, the number of unique gain values is $|\affset|$, and counting sort can be applied in $\Oh(|\affset|N)$ time. The total time complexity is thus  $\Oh((\nbits+|\affset|)N)$, which is also linear in $N$ provided that $|\affset|$ is known.
We note that counting sort on Hamming distances is also used by Lin \etal \cite{LinIJCV2016} to speed up loss-augmented inference for their NDCG surrogate loss.

\subsection{The Learning to Rank View}

Since we focus on optimizing ranking metrics, our work has connections to learning to rank \cite{learn_rank}.
Many supervised hashing formulations use loss functions defined on pairs or triplets of training examples, which correspond to \emph{pointwise} and \emph{pairwise} approaches in learning to rank terminology. 
We collectively refer to these as local ranking losses.
Since we optimize evaluation metrics defined on a ranked list, our approach falls into the \emph{listwise} category, 
and it is well-known \cite{Cao2007ICML,Wang2013listwise,Yu2014SIGIR} that listwise ranking approaches are generally superior to pointwise and pairwise approaches.

We further note that there exists a mismatch between optimizing local ranking losses and optimizing for evaluation performance.
This is because listwise evaluation metrics are \emph{position-sensitive}: errors made on individual pairs/triplets impact results differently depending on the position in the list, and more so near the top.
To address this mismatch, local ranking methods often need nontrivial weighting or sampling heuristics to focus on errors made near the top.
In fact, the sampling is especially crucial in triplet-based methods, \eg \cite{LinIJCV2016,Zhuang2016CVPR,DTSH}, since
the set of possible triplets is of size $O(N^3)$ for $N$ training examples, which can be prohibitive to enumerate.
Triplet-based methods are also popular in the metric learning literature, and it is similarly observed \cite{samplingmatters} that careful sampling and weighting are key to stable learning.
In contrast, we directly optimize listwise ranking metrics, without requiring sampling or weighting heuristics: the minibatches are sampled at random, and no weighting on training instances is used.

%
\section{Optimizing Tie-Aware Ranking Metrics}
\label{sec:optim}

In this section, we describe our approach to optimizing tie-aware ranking metrics. 
For discrete hashing, such optimization is NP-hard, since it involves combinatorial search over all configurations of binary bits.
Instead, we are interested in a relaxation approach using gradient-based deep neural networks. Therefore, we apply continuous relaxation to the discrete optimization problems.

\begin{table*}[bp]
\hrule
\vspace{.42em}
\begin{prop}
\label{prop:relax}
The continuous relaxations of $\apt$ and $\dcgt$, denoted as $\apr$ and $\dcgr$ respectively, are as follows: 
\begin{align}
 \apr(R^{(d)})& = \frac{c_d^+(c_d^+-1)}{N^+(c_d-1)} +  \frac{c_d^+}{N^+c_d}
\left[C_{d-1}^+ +1 -\frac{c_d^+-1}{c_d-1}(C_{d-1}+1)\right]
\ln \frac{C_d}{C_{d-1}}, \tag{11} \label{eq:apr}
\\
 \dcgr(R^{(d)}) & = 
\ln 2\sum_{v\in\affset}\frac{G(v)\cdv}{c_d}\int^{C_{d}+1}_{C_{d-1}+1}\frac{dt}{\ln t}.
\tag{12} \label{eq:dcgr}
\end{align}
\begin{proof}
See appendix.
\end{proof}
\end{prop}
\hrule
\end{table*}

\subsection{Continuous Relaxations}
\label{sec:optim:cont}

Our continuous relaxation needs to address two types of discrete variables.
First, as is universal in hashing formulations, the bits in the hash code are binary. 
Second, the tie-aware metrics involve integer-valued histogram bin counts $\{\ndv\}$.

We first tackle the binary bits. 
Commonly, bits in the hash code are generated by a thresholding operation using the $\sgn$ function,
\begin{align}
\Phi(x) & = (\phi_1(x),\ldots,\phi_\nbits(x)), \\ 
\phi_i(x) & = \sgn(f_i(x; w))\in\{-1,1\}, \forall i,
\label{eq:hashfunc}
\end{align}
where in our case $f_i$ are neural network activations, parameterized by $w$.
We  smoothly approximate the $\sgn$ function using the $\tanh$ function,
which is a standard technique in hashing \cite{SHK,DPSH,DTSH,Wang2015IJCAI,mihash,hashnet}: 
\begin{align}
\phi_i(x)\approx\hat{\phi}_i(x)=\tanh(\alpha f_i(x;w))\in (-1,1).
\label{eq:tanh}
\end{align}
The constant $\alpha$ is a scaling parameter.

As a result of this relaxation, both the hash mapping and the distance function \eqref{eq:hamming_dist} are now real-valued, and will be denoted $\hat{\Phi}$ and $\hat{d}_\Phi$, respectively. The remaining discreteness is from the histogram bin counts $\{\ndv\}$.
We also relax them into real-valued ``soft histograms'' $\{\cdv\}$ (described below),  whose cumulative sums are denoted $\{\Cdv\}$.
However, we face another difficulty:  the definitions of $\apt$ \eqref{eq:apt} and $\dcgt$ \eqref{eq:dcgt} both involve a finite sum with lower and upper limits $N_{d-1}+1$ and $N_d$, which themselves are variables to be relaxed.
We approximate these finite sums by continuous integrals, removing the second source of discreteness.
We outline the results in Proposition~\ref{prop:relax}, and leave proof and error analysis to the appendix.

Importantly, both relaxations  have closed-form derivatives. The differentiation for $\apr$ \eqref{eq:apr} is straightforward, while for $\dcgr$ it removes the integral in \eqref{eq:dcgr}.

\begin{figure*}[t]
\centering
\includegraphics[width=.93\linewidth]{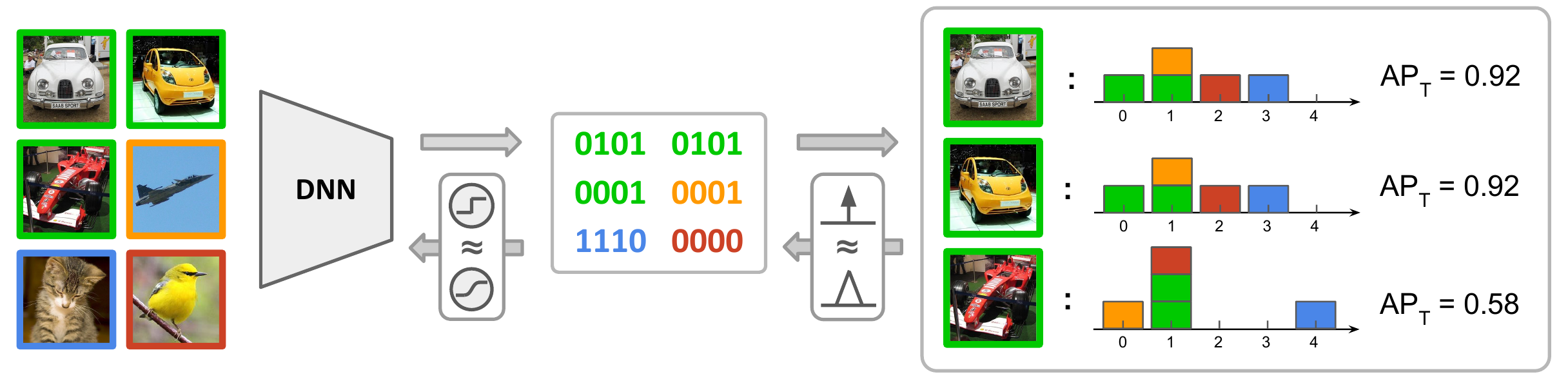}
\caption{
The flow of computation in our model. Input images are mapped to $b$-bit binary codes by a deep neural network ($b=4$ in this example).
During training, in a minibatch, each example is used as query to rank the rest of the batch, producing a histogram of Hamming distances with $(b+1)$ bins. 
Tie-aware ranking metrics ($\apt$ shown here) are computed on these histograms, and averaged over the batch.
To maintain end-to-end differentiability, we derive continuous relaxations for $\apt$ and $\ndcgt$, and employ two differentiable approximations to non-differentiable operations (backward arrows).
}
\label{fig:minibatch}
\end{figure*}

\setcounter{equation}{12}

\subsection{End-to-End Learning}
\label{sec:optim:end2end}

We perform end-to-end learning with gradient ascent.
First, as mentioned above, the continuous relaxations $\apr$ and $\dcgr$ have closed-form partial derivatives with respect to the soft histograms $\{\cdv\}$.
Next, we consider differentiating the histogram entries.
Note that before relaxation, the discrete histogram $({n}_{0,v},\ldots,{n}_{b,v})$ for $\forall v\in\affset$ is constructed as follows:
\begin{equation}
\ndv = \sum_{x_i|\aff_q(i)=v} \boldsymbol{1}[d_\Phi(x_q,x_i)=d],~ d=0,\ldots,b.
\label{eq:indicator}
\end{equation}
To relax $\ndv$ into $\cdv$, 
we employ a technique from \cite{histloss,mihash}, where the binary indicator $\boldsymbol{1}[\cdot]$ is replaced by a differentiable function $\delta(\hat{d}_\Phi(x_q,x_i),d)$ with easy-to-compute gradients.
Specifically, $\delta(\hat{d}_\Phi(x_q,x_i),d)$ linearly interpolates $\hat{d}_\Phi(x_q,x_i)$ into the $d$-th bin with slope $\Delta>0$:
\begin{align}
\forall z\in\mathbb{R}, \delta(z,d) =
\begin{cases}
1-\frac{|z - d|}{\Delta}, & |z-d| \leq \Delta, \\
0, & \text{otherwise.}
\end{cases}
\label{eq:delta}
\end{align}
Note that $\delta$ approaches the indicator function as $\Delta\to 0$.
We now have the soft histogram $\cdv$ as
\begin{align}
\cdv = \sum_{x_i|\aff_q(i)=v} \delta(\hat{d}_\Phi(x_q,x_i),d),
\end{align}
and we differentiate $\cdv$ using chain rule, \eg
\begin{align}
\frac{\partial \cdv}{\partial \hat{\Phi}(x_q)} 
& = \sum_{x_i|\aff_q(i)=v}\frac{\partial \delta(\hat{d}_\Phi(x_q,x_i),d)}{\partial \hat{d}_\Phi(x_q,x_i)}\frac{-\hat{\Phi}(x_i)}{2}.
\end{align}

The next and final step is to back-propagate gradients to the parameters of the relaxed hash mapping $\hat{\Phi}$, which amounts to differentiating the $\tanh$ function.

As shown in Fig.~\ref{fig:minibatch}, we train our models using minibatch-based stochastic gradient ascent. 
Within a minibatch, each example is retrieved against the rest of the minibatch.  That is, each example in a minibatch of size $M$ is used as the query $x_q$ once, and participates in the database for some other example $M-1$ times.
Then, the objective is averaged over the $M$ queries.

\section{Experiments}
\label{sec:exp}

\subsection{Experimental Setup}
\label{sec:exp:setup}

We conduct experiments on image retrieval datasets that are commonly used in the hashing literature: \cifar \cite{cifar10}, NUS-WIDE \cite{nuswide}, 22K \labelme \cite{labelme}, and ImageNet100 \cite{hashnet}.
Each dataset is split into a test set and a database, and examples from the database are used in training.
At test time, queries from the test set are used to perform Hamming ranking on the database, and the performance metric is averaged over the test set. 

\begin{itemize}
\vspace{-.5em}
\item \textbf{\cifar}
is a canonical benchmark for image classification and retrieval, with 60K single-labeled images from 10 classes.
Following \cite{DTSH}, we consider two experimental settings. In the first setting, the test set is constructed with 100 random images from each class (total: 1K), the rest is used as database, and 500 images per class are used for training (total: 5K). 
The second setting uses the standard 10K/50K split and the entire database is used in training.
\vspace{-.5em}
\item \textbf{NUS-WIDE}
is a multi-label dataset with 270K Flickr images. For the database, we use a subset of 196K images associated with the most frequent 21 labels as in \cite{DPSH,DTSH}.
100 images per label are sampled to construct a test set of size 2.1K, and the training set contains 500 images per label (total: 10.5K).  
\vspace{-.5em}
\item \textbf{LabelMe} is an unlabeled dataset of 22K images.
As in \cite{AdaptHash}, we randomly split LabelMe into a test test of size 2K and database of 20K. We sample 5K examples from the database for training.
\vspace{-.5em}
\item \textbf{ImageNet100} is a subset of ImageNet,  containing all the images from 100 classes. We use the same setup as in \cite{hashnet}: 130 images per class, totaling 130K images, are sampled for training, and all images in the selected classes from the ILSVRC 2012 validation set are used as queries.
\end{itemize}

Retrieval-based evaluation of supervised hashing was recently put into question by 
\cite{hashing_eval}, which points out that for multi-class datasets, binary encoding of classifier outputs is already a competitive solution. 
While this is an important point, deriving pairwise affinities from multi-class label agreement is a special case in our formulation. As mentioned in Sec.~\ref{sec:prelim}, our formulation uses a general pairwise affinity oracle $\aff$, which may or may not be derived from labels, and can be either binary or multi-level.
In fact, the datasets we consider range from multi-class/single-label (CIFAR-10, ImageNet100) to multi-label (NUS-WIDE) and unlabeled (LabelMe), and only the first case can be addressed by multi-class classification.
For multi-level affinities, we also propose a new evaluation protocol using NDCG.

We term our method {TALR} (\textbf{T}ie-\textbf{A}ware \textbf{L}earning to \textbf{R}ank), and compare it against a range of classical and state-of-the-art hashing methods.
Due to the vast hashing literature, an exhaustive comparison is unfortunately not feasible. 
Focusing on the learning to rank aspect, we select representative methods from all three categories:
\begin{itemize}
\vspace{-.25em}
\item \textbf{Pointwise (pair-based).}
Methods that define loss functions on instance pairs:
Binary Reconstructive Embeddings (BRE) \cite{BRE},
Fast Supervised Hashing (FastHash) \cite{FastHash}, 
Hashing using Auxiliary Coordinates (MACHash) \cite{affinity}, 
Deep Pair-Supervised Hashing (DPSH) \cite{DPSH},  and 
Hashing by Continuation (HashNet) \cite{hashnet}.
\vspace{-.5em}
\item  \textbf{Pairwise (triplet-based).}
We include a recent method, Deep Triplet-Supervised Hashing (DTSH) \cite{DTSH}.
\vspace{-.5em}
\item \textbf{Listwise (list-based).} We compare to two listwise ranking methods: 
Structured Hashing (StructHash) \cite{LinIJCV2016} which optimizes an NDCG surrogate, 
and Hashing with Mutual Information (MIHash) \cite{mihash} which optimizes mutual information as a ranking surrogate for binary affinities.
\end{itemize}
These selected methods include recent ones that achieve state-of-the-art results on CIFAR-10 (MIHash, DTSH), NUS-WIDE (DTSH, HashNet) and ImageNet100 (HashNet). 

Since tie-aware evaluation of Hamming ranking performance has not been reported in the hashing literature, we re-train and evaluate all methods using publicly available implementations.

\begin{table*}[t]
\centering
\begin{threeparttable}
\begin{tabular}{|l|cccc|l||cccc|l|} 
\hline
\multirow{3}{*}{\bf Method} & \multicolumn{5}{c||}{\textbf{CIFAR-10}} & \multicolumn{5}{c|}{\textbf{NUS-WIDE}} \\ 
\cline{2-11}
& 12 Bits & 24 Bits   & 32 Bits   & 48 Bits  & \multirow{9}{*}{\rotatebox{90}{S1 ($\apt$)}}
& 12 Bits & 24 Bits   & 32 Bits   & 48 Bits & \multirow{9}{*}{\rotatebox{90}{$\apt$}} 
\\
\cline{1-5} \cline{7-10}
BRE \cite{BRE}  &  0.361 & 0.448 & 0.502 & 0.533  &
&  0.561 & 0.578 & 0.589 & 0.607 &
\\ 
MACHash \cite{affinity} &  {0.628} & {0.707}  & {0.726} & {0.734} & 
&  0.361 & 0.361 & 0.361 & 0.361 &
\\ 
FastHash \cite{FastHash} &  {0.678} & {0.729}  & {0.742} & {0.757} & 
&  0.646 & 0.686 & 0.698 & 0.712 &
\\ 
StructHash \cite{LinIJCV2016} &  {0.664} & {0.693}  & {0.691} & {0.700} & 
&  0.639 & 0.645 & 0.666 & 0.669&
\\ 
DPSH \cite{DPSH}\tnote{*}  & 0.720 & 0.757 & 0.757 & 0.767 &
&  0.658  & 0.674 & 0.695 & 0.700 &
\\ 
DTSH  \cite{DTSH} & 0.725 &  0.773 & 0.781  & 0.810 & 
&  0.660  & 0.700 & 0.707 & 0.723 &
\\ 
MIHash \cite{mihash}  & 0.687 & 0.775  &  0.786 & 0.822 & 
&  0.652  & 0.693 & 0.709 & 0.723 &
\\ 
TALR-AP   & \textbf{0.732} & \textbf{0.789} & \textbf{0.800} & \textbf{0.826} &
&  \textbf{0.709} & \textbf{0.734} & \textbf{0.745} & \textbf{0.752} &
\\ 
\hline
{\bf Method} 
& {16 Bits} & {24 Bits}   & {32 Bits}   & {48 Bits} & \multirow{5}{*}{\rotatebox{90}{S2 ($\apt$) }}
& {12 Bits} & {24 Bits}   & {32 Bits}   & {48 Bits} & \multirow{5}{*}{\rotatebox{90}{AP@5K}}
\\
\cline{1-5} \cline{7-10}
DPSH \cite{DPSH}\tnote{*} &  0.908 & 0.909 & 0.917 & 0.932 &
& 0.758 & 0.793 & 0.818 & 0.830 &
\\ 
DTSH \cite{DTSH} & 0.916 &  0.924 & 0.927 & 0.934 & 
& 0.773 & 0.813 & 0.820 & 0.838 &
\\
MIHash \cite{mihash}  & 0.929 &  0.933 & 0.938 & 0.942 & 
& 0.767 & 0.784 & 0.809 & 0.834 &
\\
TALR-AP  & \textbf{0.939} & \textbf{0.941} & \textbf{0.943} & \textbf{0.945} &
&  \textbf{0.795} & \textbf{0.835} & \textbf{0.848} & \textbf{0.862}&
\\ 
\hline
\end{tabular}
\begin{tablenotes}
\small
\item[*] Trained using parameters recommended by authors of DTSH.
\end{tablenotes}
\end{threeparttable}
\vspace{-.5em}
\caption{AP comparison on CIFAR-10 and NUS-WIDE with VGG-F architecture.
On CIFAR-10, we compare all methods in the first setting (S1), and deep learning methods in the second (S2). 
We report the tie-aware $\apt$, and additionally AP@5K for NUS-WIDE.
TALR-AP optimizes tie-aware AP using stochastic gradient ascent, and achieves state-of-the-art performance. 
} 
\label{table:ap1}
\end{table*}

\subsection{AP Optimization}
\label{sec:exp:AP}

We evaluate AP optimization on the three labeled datasets, CIFAR-10, NUS-WIDE, and ImageNet100. 
As we mentioned earlier, for labeled data, affinities can be inferred from label agreements.
Specifically, in CIFAR-10 and ImageNet100, two examples are neighbors (\ie have pairwise affinity $1$) if they share the same class label. 
In the multi-labeled NUS-WIDE, two examples are neighbors if they share at least one label.

\vspace{-.5em}
\subsubsection{CIFAR-10 and NUS-WIDE}
We first carry out AP optimization experiments on the two well-studied datasets, CIFAR-10 and NUS-WIDE.
For these experiments, we perform finetuning using the ImageNet-pretrained VGG-F network \cite{vggf}, which is used in DPSH and DTSH, two recent top-performing methods.
For methods that are not amenable to end-to-end training, we train them on {fc7}-layer features from VGG-F.
On CIFAR-10, we compare all methods in the first setting, and in the second setting we compare the end-to-end methods: DPSH, DTSH, MIHash, and ours.
We do not include HashNet as it uses a different network architecture (AlexNet), but will compare to it later on ImageNet100.

We present AP optimization results  in Table~\ref{table:ap1}. 
By optimizing the relaxation of $\apt$ in an end-to-end fashion, our method (TALR-AP) achieves the new state-of-the-art in AP on both datasets, outperforming all the pair-based and triplet-based methods by significant margins.
Compared to listwise ranking solutions, TALR-AP outperforms {StructHash} significantly by taking advantage of deep learning, and outperforms MIHash by matching the training objective to the evaluation metric.
A side note is that for NUS-WIDE, it is customary in previous work \cite{DPSH,DTSH} to report AP evaluated at maximum cutoff of 5K (AP@5K), since ranking the full database is inefficient using general-purpose sorting algorithms. 
However, focusing on the top of the ranking overestimates the true AP, as seen in Table~\ref{table:ap1}. Using counting sort, we are able to evaluate $\apt$ on the full database efficiently, and TALR-AP also outperforms other methods in terms of AP@5K.

\begin{table}[t]
\centering
\begin{tabular}{|l|cccc|} 
\hline
{\bf Method} 
& {16 Bits} & {32 Bits}   & {48 Bits}   & {64 Bits} 
\\
\hline
HashNet \cite{hashnet}  & 0.5059 &  0.6306 & 0.6633 & 0.6835
\\
MIHash \cite{mihash}  & 0.5688 &  0.6608 & 0.6852 & 0.6947 
\\
TALR-AP  & \textbf{0.5892} & \textbf{0.6689} & \textbf{0.6985} & \textbf{0.7053}
\\ 
\hline
\end{tabular}
\vspace{-.5em}
\caption{AP@1000 results on ImageNet100 with AlexNet. TALR-AP outperforms state-of-the-art solutions using mutual information \cite{mihash} and continuation methods \cite{hashnet}.
} 
\label{table:ap2}
\end{table}

\vspace{-.5em}
\subsubsection{ImageNet100}
For ImageNet100 experiments, we closely follow the setup in HashNet \cite{hashnet} and fine-tune the AlexNet architecture \cite{alexnet} pretrained on ImageNet. 
Due to space limitations, we report comparisons against recent state-of-the-art methods on ImageNet100.
The first competitor is HashNet, which is empirically superior to a wide range of classical and recent methods, and was previously the state-of-the-art method on ImageNet100.
We also compare to MIHash, as it is the second-best method on CIFAR-10 and NUS-WIDE in the previous experiment. 
As in \cite{hashnet}, the minibatch size is set to 256 for all methods,  and the learning rate for the pretrained convolution and fully connected layers are scaled down,  since the model is fine-tuned on the same dataset that it was originally trained on. 
AP at cutoff 1000 (AP@1000) is used as the evaluation metric. 

ImageNet100 results are summarized in Table~\ref{table:ap2}.
TALR-AP outperforms both competing methods, and the improvement is especially significant with short hash codes (16 and 32 bits). 
This indicates that our direct optimization approach produces better compact binary representations that preserve desired rankings. 
The state-of-the-art performance with compact codes has important implications for cases where memory and storage resources are restricted (\eg mobile applications), and for indexing large-scale databases. 

\begin{table*}[t]
\centering
\begin{threeparttable}
\begin{tabular}{|l|cccc||cccc|} 
\hline
 \multirow{2}{*}{\bf Method} & \multicolumn{4}{c||}{\textbf{NUS-WIDE}} & \multicolumn{4}{c|}{\textbf{LabelMe}} \\ 
\cline{2-9}
 &  {16 Bits} & {32 Bits}   & {48 Bits}   & {64 Bits} 
 & {16 Bits} & {32 Bits}   & {48 Bits}   & {64 Bits}\\
\hline
BRE \cite{BRE}\tnote{*} & 0.805 & 0.817 & 0.827 & 0.834
&  0.807 & 0.848 & 0.871 & 0.880
\\
MACHash \cite{affinity} & 0.821 & 0.821 & 0.821 & 0.821
&  0.683 & 0.683 & 0.683 & 0.687
\\ 
FastHash \cite{FastHash}  & {0.885} & {0.896}  & {0.899} & {0.902} 
&  0.844 & 0.868 & 0.855 & 0.864
\\ 
DPSH \cite{DPSH}  & 0.895 & 0.905 & 0.909 & 0.909 
&  0.844 & 0.856 & 0.871 & 0.874
\\ 
DTSH \cite{DTSH}  & 0.896 &  0.905 & 0.911  & 0.913  
&  0.838 & 0.852 & 0.859 & 0.862
\\ 
StructHash \cite{LinIJCV2016} & 0.889 & 0.893 & 0.894 & 0.898
&  0.857 & {0.888} & 0.904 & 0.915
\\
MIHash \cite{mihash}  & 0.886 &  0.903 & 0.909  & 0.912 
&  0.860 & 0.889 & 0.907 & 0.914
\\ 
TALR-NDCG   & \textbf{0.903} & \textbf{0.910} & \textbf{0.916} & \textbf{0.927} 
&  \textbf{0.866} & \textbf{0.895} & \textbf{0.908} & \textbf{0.917}
\\ 
\hline
\end{tabular}
\begin{tablenotes}
\small
\item[*] Evaluated on the the 5K training subset due to kernel-based formulation.
\end{tablenotes}
\end{threeparttable}
\vspace{-.5em}
\caption{NDCG comparison on NUS-WIDE (VGG-F architecture) and LabelMe (shallow models on GIST features). 
TALR-NDCG optimizes tie-aware NDCG using stochastic gradient ascent,
and consistently outperforms competing methods. 
} 
\label{table:ndcg}
\end{table*}

\subsection{NDCG Optimization}
\label{sec:exp:NDCG}

We evaluate NDCG optimization with a multi-level affinity setup, \ie the set of affinity values $\affset$ is a finite set of non-negative integers.
Multi-level affinities are common in information retrieval tasks, and offer more fine-grained specification of the desired structure of the learned Hamming space.
To our knowledge, this setup has not been considered in the hashing literature. 

In the multi-label NUS-WIDE dataset, we define the affinity value between two examples as the number of labels they share, and keep other settings the same as in the AP experiment.
For the unlabeled LabelMe dataset, we derive affinities by thresholding the Euclidean distances between examples. 
Inspired by an existing binary affinity setup \cite{AdaptHash} that defines neighbors as having Euclidean distance within the top $5\%$ on the training set, 
we use four thresholds $\{5\%, 1\%, 0.2\%, 0.1\%\}$ and assign affinity values $\{1, 2, 5, 10\}$. 
This emphasizes assigning high ranks to the closest neighbors in the original feature space.
We learn shallow models on precomputed GIST features on LabelMe. For gradient-based methods, this means using linear hash functions, \ie $f_i(x;w)=w_i^\top x$, in \eqref{eq:hashfunc}.
For methods that are not designed to use multi-level affinities (FastHash, MACHash, DPSH, MIHash), we convert the affinities into binary values; this reduces to the standard binary affinity setup on both datasets.

We give NDCG results in Table~\ref{table:ndcg}.
Again, our method with the tie-aware NDCG objective (TALR-NDCG) outperforms all competing methods on both datasets.
Interestingly, on LabelMe where all methods are restricted to learn shallow models on GIST features, we observe slightly different trends compared to other datasets.
For example, without learning deep representations, DPSH and DTSH appear to perform less competitively, indicating a mismatch between their objectives and the evaluation metric.
The closest competitors to TALR-NDCG on LabelMe are indeed the two listwise ranking methods: StructHash which optimizes a NDCG surrogate using boosted decision trees, and MIHash which is designed for binary affinities.
TALR-NDCG outperforms both methods, and notably does so with linear hash functions, which have lower learning capacity compared StructHash's boosted decision trees. This highlights the benefit of our direct optimization formulation.

\begin{figure*}[ht]
\centering
\includegraphics[width=\linewidth]{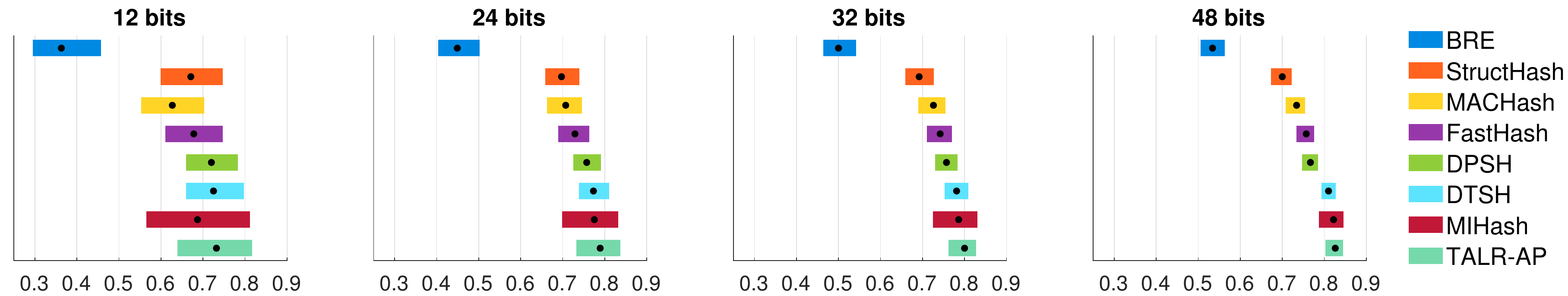}
\vspace{-1.5em}
\caption{
Effects of tie-breaking: we plot the ranges of test-time mAP values spanned by all possible tie-breaking strategies, for all methods considered in the CIFAR-10 experiment (first setting).
Horizontal axis: mAP. Black dots: values of tie-aware $\apt$.
Without controlling for tie-breaking, relative performance comparison between different methods can be ambiguous. The ambiguity is eliminated by tie-awareness.
}
\label{fig:ties}
\vspace{-.2em}
\end{figure*}

\subsection{Effects of Tie-Breaking}
\label{sec:exp:ties}
We lastly discuss the effect of tie-breaking in evaluating hashing algorithms.
As mentioned in Sec.~\ref{sec:ties}, tie-breaking is an uncontrolled parameter in current evaluation protocols, which can affect results, and even be exploited.
To demonstrate this, we consider for example the AP experiment in CIFAR-10's first setting, presented in Sec.~\ref{sec:exp:AP}. 
For each method included in this experiment, we plot the range of test set mAP spanned by all possible tie-breaking strategies. 
As can be seen in Fig.~\ref{fig:ties},  the ranges corresponding to different methods generally overlap; therefore, without controlling for tie-breaking, relative performance comparison between different methods is essentially ambiguous. 
The ranges shrink as code length increases, since the number of ties generally decreases when there are more bins in the histogram. 

Current hashing methods usually compute test-time AP and NDCG using random tie-breaking and general-purpose sorting algorithms.
Interestingly, in all of our experiments, we observe that this produces values very close to the tie-aware $\apt$ and $\ndcgt$.
The reason is that with a randomly ordered database, averaging the tie-unaware metric over a sufficiently large test set behaves similarly to the tie-aware solution of averaging over all permutations.
Therefore, the results reported in the current literature are indeed quite fair,  and so far we have found no evidence of exploitation of tie-breaking strategies.
Still,  we recommend using tie-aware ranking metrics in evaluation, as they completely eliminate ambiguity, and counting sort on Hamming distances is much more efficient than general-purpose sorting.

We note that although random tie-breaking is an approximation to tie-awareness at \emph{test time}, it does not answer the question of how to \emph{optimize} the ranking metrics during training.
Our original motivation is to optimize ranking metrics for hashing, and the existence of closed-form tie-aware ranking metrics is what makes direct optimization feasible.

\section{Conclusion}
\label{sec:conclusion}

We have proposed a new approach to hashing for nearest neighbor retrieval, with an emphasis on directly optimizing evaluation metrics used at test-time.
A study into the commonly used retrieval by Hamming ranking setup led us to consider the issue of ties, 
and we advocate for using tie-aware versions of ranking metrics. 
We then make the novel contribution of optimizing tie-aware ranking metrics for hashing, focusing on the important special cases of AP and NDCG.
To tackle the resulting discrete and NP-hard optimization problems, we derive their continuous relaxations. 
Then, we perform end-to-end stochastic gradient ascent with deep neural networks.
This results in the new state-of-the-art for common image retrieval benchmarks.

\section*{Acknowledgements}
The authors would like to thank Qinxun Bai, Peter Gacs, and Dora Erdos for helpful discussions. 
This work is supported in part by a BU IGNITION award, NSF grant 1029430, and gifts from Nvidia.

{
\small
\bibliographystyle{plain}
\bibliography{bibs}
}

\onecolumn
\appendix
\section*{Appendix}

\section{Proof of Proposition 1}
\label{appendix:prop1}
\begin{proof}
Our proof essentially restates the results in \cite{ties} using our notation. 
In \cite{ties}, a \emph{tie-vector} $T=(t_0,\ldots,t_{d+1})$ is defined, where $t_0=0$ and the next elements indicate the ending indices of the equivalence classes in the ranking, \eg $t_1$ is the ending index of $R^{(0)}$, and so on.
Using our notation, we can see that $R^{(d)}=(R_{1+t_d},\ldots,R_{t_{d+1}})$, and $t_d=N_{d-1}=\sum_{j=0}^{d-1}n_j$.

We first consider $\apt$. In Section~2.4 of \cite{ties}, the tie-aware AP at cutoff $k$ is defined as
\begin{align}
\apt@k(R) = \frac{\sum_{j=1}^k\frac{n_i^+}{n_i}\left(N_{i-1}^++(j-t_i-1)\frac{n_i^+-1}{n_i-1}+1\right)\frac{1}{j}}
{\sum_{j=1}^{|S|}\aff_q(j)},
\end{align}
where $i$ is the index of the tie that item $j$ is in.
To derive $\apt$ in our formulation, we take $k$ to be the maximum possible cutoff $|S|$, and substitute by definition $N^+=\sum_{j=1}^{|S|}\aff_q(j),t_i=N_{i-1}$:
\begin{align}
\apt(R) & =\frac{1}{N^+}\sum_{j=1}^{|S|}\frac{\frac{n_i^+}{n_i}\left(N_{i-1}^++(j-N_{i-1}-1)\frac{n_i^+-1}{n_i-1}+1\right)}{j}.
\end{align}
It is clear that this sum decomposes additively over $j$. Therefore, we can explicitly compute the contribution from items in a single tie $\Rd$,
\begin{align}
\apt(\Rd) & =\frac{1}{N^+}\sum_{j=N_{d-1}+1}^{N_d}\frac{\frac{n_d^+}{n_d}\left(N_{d-1}^++(j-N_{d-1}-1)\frac{n_d^+-1}{n_d-1}+1\right)}{j},
\end{align}
and this gives \eqref{eq:apt}.

Next, tie-aware DCG is given in Section~2.6 of \cite{ties} as 
\begin{align}
\dcgt@k(R)=\sum_{d}\left[
\left(\frac{1}{n_d}\sum_{j=t_d+1}^{t_{d+1}} G(\aff_q(j))\right)
\sum_{j=t_d+1}^{\min(t_{d+1},k)}D(j)
\right].
\end{align}
Again, we consider a single tie $\Rd$, take $k=|S|$, and make the substitution $t_d=N_{d-1}$:
\begin{align}
\dcgt(\Rd) & =\frac{1}{n_d}\sum_{j\in\Rd}G(\aff_q(j))\sum_{j=N_{d-1}+1}^{N_d}D(j)\\
& =\frac{1}{n_d}\sum_{v\in\affset}\overbrace{\sum_{j\in \Rd}\boldsymbol{1}[v=\aff_q(j)]}^{\ndv}G(v)\sum_{j=N_{d-1}+1}^{N_d}D(j)\\
& =\frac{1}{n_d}\sum_{v\in\affset}G(v)\ndv\sum_{j=N_{d-1}+1}^{N_d}D(j).
\end{align}
This completes the derivation for \eqref{eq:dcgt}.
\end{proof}

\section{Proof of Proposition 2}
\label{appendix:prop2}
\begin{proof}
First, we denote the summand in the definition of $\apt$ \eqref{eq:apt} as $\beta_d(t)$, and rewrite it  as
\begin{align}
\beta_d(t) =\frac{N_{d-1}^++(t-N_{d-1}-1)\frac{n_d^+-1}{n_d-1}+1}{t} 
= \frac{n_d^+-1}{n_d-1}+\frac{N_{d-1}^+ +1 -\frac{n_d^+-1}{n_d-1}(N_{d-1}+1)}{t}.
\end{align}
It is of the form $A+B/t$ where  $A,B$ are constant in $t$.
We proceed with the summation  over $t$ in \eqref{eq:apt}:
\begin{align}
\apt(\Rd) 
& = \frac{n_d^+}{n_dN^+}\sum_{t=N_{d-1}+1}^{N_d}\beta_d(t) \\
& = \frac{n_d^+}{n_dN^+} \left[  
\frac{n_d^+-1}{n_d-1}n_d + \left(N_{d-1}^+ +1 -\frac{n_d^+-1}{n_d-1}(N_{d-1}+1)\right)\sum_{t=N_{d-1}+1}^{N_d}\frac{1}{t}
\right].
\label{eq:harmonic}
\end{align}

The main obstacle in the continuous relaxation is the finite sum in \eqref{eq:harmonic}, which has in its limits $N_{d-1}$ and $N_d$, variables to be relaxed.
However, it is a partial sum of the harmonic series, which can be well approximated  
by differences of the natural logarithm:
\begin{align}
\sum_{t=N_{d-1}+1}^{N_d}\frac{1}{t} \approx
\int_{N_{d-1}}^{N_d}\frac{dt}{t}=\ln(N_d)-\ln(N_{d-1}).
\label{eq:ap-approx}
\end{align}
In fact, \eqref{eq:ap-approx} corresponds to the \emph{midpoint rule} in approximating definite integrals by finite sums, but is applied in the reverse direction.
The relaxation of $\apt$ is then derived as follows:
\begin{align}
\apt(\Rd) & \approx
\frac{n_d^+}{n_dN^+} \left[  
\frac{n_d^+-1}{n_d-1}n_d + \left(N_{d-1}^+ +1 -\frac{n_d^+-1}{n_d-1}(N_{d-1}+1)\right)
\ln\frac{N_d}{N_{d-1}} \right]\\
\Rightarrow ~ \apr(\Rd) & = 
\frac{c_d^+(c_d^+-1)}{(c_d-1)N^+} + \frac{c_d^+}{c_dN^+}
\left[C_{d-1}^+ +1 -\frac{c_d^+-1}{c_d-1}(C_{d-1}+1)\right]
\ln \frac{C_d}{C_{d-1}}.
\end{align}
Note that $N^+=\sum_dn_d^+$ is a constant for a fixed query and fixed database, thus it is not affected by the relaxation.

Next, we consider $\dcgt$, where the sum of logarithmic\footnote{Other types of discounts are also used in the literature, including linear discount: $D(t)\propto \frac{1}{t}$. It is easy to see that our technique also applies.} discount values similarly involves variables to be relaxed in its limits.
Thus, the same approximation strategy using continuous integrals is used.
\begin{align}
\sum_{t=N_{d-1}+1}^{N_d}\!\!D(t)
= \sum_{t=N_{d-1}+1}^{N_d}\frac{1}{\log_2(t+1)}
\approx \int_{N_{d-1}}^{N_d}\frac{dt}{\log_2(t+1)}
= \ln2 \int_{N_{d-1}+1}^{N_d+1}\frac{dt}{\ln t}.
\end{align}
Combining with the definition  in \eqref{eq:dcgt}, we get the continuous relaxation of $\dcgt$:
\begin{align}
\dcgr(R^{(d)}) & =  
\ln 2\sum_{v\in\affset}\frac{G(v)\cdv}{c_d}\int_{C_{d-1}+1}^{C_d+1}\frac{dt}{\ln t}\\
& = \ln 2\sum_{v\in\affset}\frac{G(v)\cdv}{c_d}\left[Li(C_d+1)-Li(C_{d-1}+1)\right]
\end{align}
where $Li$ is the logarithmic integral function: $Li(x)=\int_0^x\frac{dx}{\ln x}$.

\end{proof}

\section{Approximation Error Analysis}
We now analyze the approximation error when doing the continuous relaxations.
We take  $\apt$ as example, and note that the analysis for $\dcgt$ is similar.

The continuous relaxation for $\apt(\Rd)$ is given in \eqref{eq:ap-approx}, which replaces a finite sum with a defnite integral, where the finite sum has $N_d-N_{d-1}=n_d$ summands. 
First, we consider the case  where there are no ties, or $n_d\in\{0,1\}$, \ie the $d$-th histogram bin is either empty or contains a single item.
In this case, we can directly evaluate the lefthand side sum  in \eqref{eq:ap-approx} to be either $0$ or $\frac{1}{N_d}$, without using the integral approximation. 
Therefore, when there are no ties, there is no approximation error.

Next we consider $n_d\geq 2$. Let the $N$-th harmonic number be $H(N)=\sum_{t=1}^{N} \frac{1}{t}$, then the lefthand side of \eqref{eq:ap-approx} is  exactly $H(N_d)-H(N_{d-1})$.
It is well known that the harmonic number can be closely approximated as
\begin{align}
H(N) = \gamma + \ln(N) + \frac{1}{2N}+ O\left(\frac{1}{12N^2}\right), 
\end{align} 
where $\gamma\approx 0.5772$ is Euler’s constant. 
A direct application of this approximation gives the following:
\begin{align}
H(N_d)&=\gamma + \ln(N_d) + \frac{1}{2N_d} + O\left(\frac{1}{12N_d^2}\right)\\ 
H(N_{d-1})&=\gamma + \ln(N_{d-1}) + \frac{1}{2N_{d-1}} + O\left(\frac{1}{12N_{d-1}^2}\right)\\ 
\Rightarrow H(N_d)-H(N_{d-1})&=\ln(N_d)-\ln(N_{d-1})+O\left(\frac{1}{2N_{d-1}}-\frac{1}{2N_{d}}\right).
\label{eq:ap-approx-err}
\end{align}
Comparing \eqref{eq:ap-approx-err} with \eqref{eq:ap-approx}, we see that the approximation error is 
\begin{align}
O\left(\frac{1}{2N_{d-1}}-\frac{1}{2N_{d}}\right) = O\left(\frac{n_d}{2N_{d-1}N_{d}}\right) 
= O\left(\frac{n_d}{2N_{d-1}(N_{d-1}+n_d)}\right) 
= O\left(\frac{n_d}{2N_{d-1}^2}\right).
\end{align}
The error is proportional to $n_d$, the number of items in the $d$-th bin in the Hamming distance histogram. 
However, even if $n_d$ is large, the error is in general still small, since it has $N_{d-1}^2$ in the denominator.
Note that \eqref{eq:ap-approx} can be further tightened by including the $\frac{1}{2N}$  term, or even higher order terms in the approximation of Harmonic numbers, but the approximation using the first two terms (Euler's constant and natural log) is already quite tight, and is in fact used widely.

\section{Tightening the $\tanh$ Relaxation}
As is common among relaxation-based hashing methods, we relax the binary constraints by replacing the discontinuous $\sgn$ function with $\tanh$ \eqref{eq:tanh}.
With this  simple relaxation, the performance gains are mainly due to optimizing the proposed objectives. 
Nevertheless, it is conceivable that the continuation method, \eg as employed by HashNet \cite{hashnet}, can tighten the relaxation and lead to better results. 
To provide a concrete example, below we report improved results for TALR-AP on CIFAR-10 (setting 1), when we increase the scaling $\alpha$ in $\tanh$ over time, instead of using a fixed value.

\begin{table}[h]
\centering
\begin{tabular}{|c|c|c|c|c|}
\hline
$\apt$ & 12 bits & 24 bits & 32 bits & 48 bits \\
\hline
$\alpha=40$ &  0.732 & 0.789 & 0.800 & 0.826 \\
$\alpha\rightarrow \infty$ & 0.751 & 0.804 & 0.813  & 0.830 \\
\hline
\end{tabular}
\end{table}

\newcommand{\bigPhi}{\boldsymbol{\hat{\Phi}}}
\newcommand{\D}{\mathcal{D}}
\newcommand{\vp}{\varoplus}
\newcommand{\cdvi}{c_{d,v}^{(i)}}
\newcommand{\advi}{\alpha_{d,v}^{(i)}}
\newcommand{\bdvi}{\boldsymbol{\beta}_{d,v}^{(i)}}

\section{Efficient Minibatch Backpropagation}
As mentioned in Sec.~\ref{sec:optim:end2end}, our models are trained using minibatch SGD.
To maximally utilize supervision, we use the following strategy: each example in the minibatch is used to query the rest of the batch (which acts as the database), and the resulting objective values are averaged. 
Here, we detail the derivation of the backpropagation rules according to this formulation.
We take inspirations from \cite{cakir2018hashing} and \cite{fewshotAP}, both of which similarly average listwise objectives  designed for binary affinities.
Our technique can be seen as a direct generalization of \cite{cakir2018hashing} in that we support multi-level affinities.


Consider a minibatch of size $M$, $\{x_1,\ldots,x_M\}$. 
We use a unified shorthand $\mathcal{O}^{(i)}$ to denote the (relaxed) objective value when $x_i$ is the query, which can either be $\apr$ or $\dcgr$ in our formulation.
The overall minibatch objective is then $\mathcal{O}=\frac{1}{M} \sum_{i=1}^M\mathcal{O}^{(i)}$.
For the entire minibatch, we group the relaxed hash mapping output into a $b\times M$ matrix,
\begin{align}
\bigPhi =  \left[
\hat{\Phi}(x_1) ~ \hat{\Phi}(x_2) ~ \cdots ~ \hat{\Phi}(x_M)
\right]\in\mathbb{R}^{b\times M}.
\end{align}
We consider the multi-level affinity setup where affinity values are from a finite set $\affset$, which includes binary affinities as a special case, \ie when $\affset=\{0,1\}$.
The Jacobian of the minibatch objective with respect to $\bigPhi$ can be written as
\begin{align}
\frac{\partial \mathcal{O}}{\partial {\bigPhi}} 
 = \frac{1}{M} \sum_{i=1}^M\frac{\partial \mathcal{O}^{(i)}}{\partial {\bigPhi}}
 = \frac{1}{M}\sum_{i=1}^M\sum_{d=0}^b\sum_{v\in\affset}
\frac{\partial \mathcal{O}^{(i)}}{\partial \cdvi} \frac{\partial \cdvi}{\partial {\bigPhi}},
\label{eq:grad_minibatch}
\end{align}
where as defined earlier in Sec.~\ref{sec:optim:cont}, $\cdvi$ is the continuous relaxation of $\ndv^{(i)}$, the $d$-th bin in the distance histogram conditioned on affinity level $v$.
The superscript $^{(i)}$ indicates that the current query is $x_i$ and the database is $\{x_1,\ldots,x_M\}\setminus \{x_i\}$.

Evaluating this Jacobian involves two steps. 
First, we need to compute the partial derivative $\partial \mathcal{O}^{(i)}/\partial \cdvi,\forall d\forall v$.
Note that this is exactly the differentiation of $\apr$ and $\dcgr$, and as we pointed out in Sec.~\ref{sec:optim:cont}, both can be evaluated in  closed form.
We use variables $\alpha$ to denote these partial derivatives,  deferring the details of derivation:
\begin{align}
\alpha_{d,v}^{(i)}=\frac{\partial \mathcal{O}^{(i)}}{\partial \cdvi}.
\label{eq:alpha_dv}
\end{align}

Next,  we need to evaluate the Jacobian $\partial \cdvi/\partial \bigPhi$, which is essentially differentiating the soft histogram binning process.
Let us  consider each column of this Jacobian. 
First, for $\forall j\neq i$, using chain rule,
\begin{align}
\frac{\partial \cdvi}{\partial \hat{\Phi}(x_j)} 
 = \frac{\partial \cdvi}{\partial \hat{d}_\Phi(x_i,x_j)} \frac{\partial \hat{d}_\Phi(x_i,x_j)}{\partial \hat{\Phi}(x_j)}
& = \boldsymbol{1}[\aff_i(j)=v]\frac{\partial \delta(\hat{d}_\Phi(x_i,x_j),d)}{\partial \hat{d}_\Phi(x_i,x_j)} \frac{\partial \hat{d}_\Phi(x_i,x_j)}{\partial \hat{\Phi}(x_j)}\\
& = \boldsymbol{1}[\aff_i(j)=v]\delta_d'(\hat{d}_\Phi(x_i,x_j))\frac{-\hat{\Phi}(x_i)}{2}\\
& \overset{\Delta}{=} \beta_{d,v}(i,j)\frac{-\hat{\Phi}(x_i)}{2},
\end{align}
where $\delta_d'$ is the derivative of the single-argument function $\delta(\cdot,d)$ \eqref{eq:delta}, and we define the shorthand
\begin{align}
\beta_{d,v}(i,j) & = \boldsymbol{1}[\aff_i(j)=v] \delta_d'(\hat{d}_\Phi(x_i,x_j)).
\label{eq:beta}
\end{align}
Note that $\beta$ is symmetric, \ie $\beta_{d,v}(i,j)=\beta_{d,v}(j,i)$, which follows from the fact that both the affinity $\aff$ and the distance function $\hat{d}_\Phi$ are symmetric.

For $j=i$, we have a similar result:
\begin{align}
\frac{\partial \cdvi}{\partial \hat{\Phi}(x_i)}  = 
\sum_{k\neq i}\frac{\partial \cdvi}{\partial \hat{d}_\Phi(x_i,x_k)} \frac{\partial \hat{d}_\Phi(x_i,x_k)}{\partial \hat{\Phi}(x_i)}  
 = \sum_{k\neq i} \beta_{d,v}(i,k)\frac{-\hat{\Phi}(x_k)}{2} .
\end{align}
To unify these two cases, we require that $\beta_{d,v}(i,i)\equiv 0,\forall i$. We now have a unified expression for the $j$-th column in the Jacobian $\partial \cdvi/\partial \bigPhi$:
\begin{align}
\frac{\partial \cdvi}{\partial \hat{\Phi}(x_j)} = 
-\frac{1}{2}\beta_{d,v}(i,j)\hat{\Phi}(x_i)
- \frac{\boldsymbol{1}[j=i]}{2} \sum_{k=1}^M \beta_{d,v}(i,k)\hat{\Phi}(x_k).
\end{align}

We now obtain a compact matrix form for $\partial \cdvi/\partial \bigPhi$. 
Let $\bdvi=(\beta_{d,v}(i,1),\ldots,\beta_{d,v}(i,M))\in\mathbb{R}^M$.
Also, let $\boldsymbol{e}_i$ be the $i$-th standard basis vector in $\mathbb{R}^M$, \ie the $i$-th element is $1$ and all others are $0$.
We have that
\begin{align}
\frac{\partial \cdvi}{\partial \bigPhi} & = 
-\frac{1}{2}\hat{\Phi}(x_i)(\bdvi)^\top
-\left[\sum_{k=1}^M\frac{1}{2}\beta_{d,v}(i,k)\hat{\Phi}(x_k)\right]\boldsymbol{e}_i^\top
 = 
-\frac{1}{2}\left[\hat{\Phi}(x_i)(\bdvi)^\top
+ \bigPhi \bdvi\boldsymbol{e}_i^\top\right].
\end{align}

Finally, we complete \eqref{eq:grad_minibatch} using the result above.
The main trick is to change the ordering of sums: we bring the sum over $i=1,\ldots,M$ inside,
\begin{align}
\frac{\partial\mathcal{O}}{\partial\bigPhi} 
& = \frac{1}{M}\sum_{i=1}^M\sum_{d=0}^b\sum_{v\in\affset}
\frac{\partial \mathcal{O}_i}{\partial \cdvi} \frac{\partial \cdvi}{\partial \bigPhi }\\
 &=  \frac{1}{M}\sum_{d=0}^b\sum_{v\in\affset}\sum_{i=1}^M -\frac{1}{2}\advi
\left[ \hat{\Phi}(x_i)(\bdvi)^\top
+ \bigPhi \bdvi \boldsymbol{e}_i^\top \right]\\
 &= - \frac{1}{2M}\sum_{l=0}^b\sum_{v\in\affset}\left[
\sum_{i=1}^M\advi\hat{\Phi}(x_i)(\bdvi)^\top
+ \bigPhi \sum_{i=1}^M\advi\bdvi\boldsymbol{e}_i^\top \right].
\label{eq:intermediate}
\end{align}
To further simplify this result, we define  two $M\times M$ matrices:
\begin{align}
A_{d,v} & =\mathrm{diag} (\alpha_{d,v}^{(1)}~,\ldots,\alpha_{d,v}^{(M)})\in\mathbb{R}^{M\times M},\label{eq:A} \\
B_{d,v} & =\left[ \boldsymbol{\beta}_{d,v}^{(1)} ~ \cdots ~ \boldsymbol{\beta}_{d,v}^{(M)}\right]
= \left[\begin{matrix}
\beta_{d,v}(1,1) & \beta_{d,v}(2,1) & \cdots & \beta_{d,v}(M,1) \\ 
\beta_{d,v}(1,2) & \beta_{d,v}(2,2) &\cdots & \beta_{d,v}(M,2) \\ 
\vdots & \vdots & \ddots & \vdots  \\ 
\beta_{d,v}(1,M) & \beta_{d,v}(2,M) &\cdots & \beta_{d,v}(M,M) \\ 
\end{matrix} \right]
\in \mathbb{R}^{M\times M}.
\label{eq:B}
\end{align}
Then, we arrive at the following simplification of \eqref{eq:intermediate} and \eqref{eq:grad_minibatch},
\begin{align}
\frac{\partial \mathcal{O}}{\partial {\bigPhi}} 
= -\frac{1}{2M}\sum_{d=0}^b\sum_{v\in\affset}\left[ \bigPhi A_{d,v} (B_{d,v})^\top  +\bigPhi B_{d,v} A_{d,v} \right] 
=  -\frac{\bigPhi}{2M}\sum_{d=0}^b\sum_{v\in\affset}\left(  A_{d,v}B_{d,v}  + B_{d,v} A_{d,v} \right).
\end{align}
Note that we have used the fact that $B_{d,v}$ is a symmetric matrix \eqref{eq:B}, which is because $\beta$ is symmetric, as mentioned earlier.
This operation can be implemented efficiently using only matrix multiplications and additions.
Also, since $A_{d,v}$ is a diagonal matrix, multiplying it with $B_{d,v}$ essentially scales the rows or columns of $B_{d,v}$, which is an $O(M^2)$ operation as opposed to $O(M^3)$ as in general matrix multiplication.
The entire time complexity is therefore $O(b|\affset|M^2)$.

At this point, we have completed the differentiation of the minibatch objective $\mathcal{O}$ with respect to the relaxed hash mapping output, $\bigPhi$.
Further backpropagation is straightforward, since $\bigPhi$ is obtained by applying a pointwise $\tanh$ function on the raw activations from the previous layer.

\section{Implementation Details}
\label{sec:TALR:details}

We have mentioned that the gradients of the continuous relaxations to the tie-aware ranking metrics can be evaluated in closed form. 
This fact is important for performing gradient ascent.
However, it can be seen that the continuous relaxations are quite complicated, thus deriving and implementing the gradients by hand can be tedious.
Also, automatic differentiation tools can only offer limited help in this case.

Below, we present simplified versions of the continuous objectives, that are much easier to derive and implement.
Specifically, for $\apr$ we give an inexact approximation, and for $\dcgr$ we give a lower bound.
Empirically, optimizing the simplified versions gives performances that are very similar to the optimizing the original continuous relaxations.


\subsection{A Simplified Version of Tie-Aware AP}
\label{sec:TALR:details:AP}
We repeat the definition of $\apr$ below:
\begin{align}
\apr(\Rd) & =  \frac{c_d^+(c_d^+-1)}{(c_d-1)N^+} + \frac{c_d^+}{c_dN^+}
\left[C_{d-1}^+ +1 -\frac{c_d^+-1}{c_d-1}(C_{d-1}+1)\right]
\ln \frac{C_d}{C_{d-1}}.
\end{align}
Suppose that we want to differentiate $\apr$ with respect to some histogram bin $c_d^+$.
The exact partial derivative can be written as:
\begin{align}
& \sum_{d=0}^b\frac{\partial \apr(R^{(d)})}{\partial c_d^{+}}\\
= &\sum_{d=0}^b\frac{\partial}{\partial c_d^+}\left\{
\frac{c_d^+(c_d^+-1)}{(c_d-1)N^+} + \frac{c_d^+}{c_dN^+}
\left[C_{d-1}^+ +1 -\frac{c_d^+-1}{c_d-1}(C_{d-1}+1)\right]
\ln \frac{C_d}{C_{d-1}}
\right\}.
\end{align}
Again, computing this can be tedious and error-prone.

To derive an inexact version with a simpler form, we first do a change of variables in the definition of $\apt$ \eqref{eq:apt} and rewrite it as
\begin{align}
\apt(R^{(d)}) & = \frac{n_d^+}{N^+n_d}\sum_{j=1}^{n_d}
\frac{N_{d-1}^++1+(j-1)\frac{n_d^+-1}{n_d-1}}{N_{d-1}+j} \label{eq:apt-alt}\\
& \define \frac{1}{N^+}\frac{n_d^+}{n_d}\sum_{j=1}^{n_d}\eta_d(j).
\end{align}
Then, we simply replace the sum $\sum_{j=1}^{n_d}\eta_d(j)$ by repeating its midpoint: $\sum_{j=1}^{n_d}\eta_d(j)\approx n_d\eta_d(\frac{n_d+1}{2})$.
Then, the simplified version is derived as
\begin{align}
\apt(R^{(d)}) & \approx \frac{1}{N^+}\frac{n_d^+}{n_d} \cdot
n_d\frac{N_{d-1}^++1+\frac{n_d-1}{2}\frac{n_d^+-1}{n_d-1}}{N_{d-1}+\frac{n_d+1}{2}}\\
& = \frac{n_d^+}{N^+}\cdot  \frac{2N_{d-1}^++n_d^++1}{2N_{d-1}+n_d+1}\\
& = \frac{n_d^+}{N^+}\cdot \frac{N_{d-1}^++N_d^++1}{N_{d-1}+N_d+1}.
\label{eq:ap_s}
\end{align}

The simplified   continuous relaxation, which we name  $\mathrm{AP}_s$, is now as follows:
\begin{align}
\mathrm{AP}_s(\Rd)= \frac{c_d^+}{N^+}\cdot \frac{C_{d-1}^++C_d^++1}{C_{d-1}+C_d+1}.
\end{align}
Deriving the closed-form gradients for this simplification is much less involved.

\subsection{A Lower Bound for Tie-Aware DCG}
\label{sec:TALR:details:DCG}
Similar to the case of tie-aware AP, we also derive a simplified version for the tie-aware DCG. 
Here, we will actually derive a lower bound for $\dcgt$ and then continuously relax it. 
Maximizing this lower bound then serves as a surrogate for  maximizing $\dcgt$.

First we revisit the definition of $\dcgt$, plugging in the actual forms of the gain and discount functions:
\begin{align}
\dcgt(\Rd) 
&=\sum_{v\in\affset} \frac{(2^v-1)\ndv}{n_d}\sum_{t=N_{d-1}+1}^{N_d}\frac{1}{\log_2(t+1)}.
\end{align}
Note that the function $1/\log_2(\cdot)$ is a convex function. Therefore we can use Jensen's inequality to lower bound the second sum:
\begin{align}
\dcgt(\Rd)& \geq\sum_{v\in\affset} \frac{(2^v-1)\ndv}{n_d}\cdot n_d \cdot \frac{1}{\log_2(N_{d-1}+\frac{n_d+1}{2}+1)}\\
&=  \frac{\sum_{v\in\affset}(2^v-1)\ndv}{\log_2(N_{d-1}+\frac{1}{2}n_d+\frac{3}{2})}.
\end{align}
And now we can see that the continuous relaxation of the lower bound, denoted as $\text{DCG}_s$, should be
\begin{align}
\text{DCG}_s(\Rd)=  \frac{\sum_{v\in\affset}(2^v-1)\cdv}{\log_2(C_{d-1}+\frac{1}{2}c_d+\frac{3}{2})}.
\label{eq:dcg_s}
\end{align}

\subsection{Differentiating the Objectives}
\label{sec:TALR:details:matrix}
Earlier we have deferred the details of computing \eqref{eq:alpha_dv}, the partial derivatives of the minibatch objective.
We now complete our derivations.
Without loss of generality, we use $\mathcal{O}$ to denote the objective (relaxation of AP or NDCG), and assume multi-level affinity, indexed by $v$.
Recall that the set of affinity values is denoted $\affset$;
given a query and a database, we can  thus build $|\affset|$ soft histograms, $\boldsymbol{c}_v=(c_{0,v},\ldots,c_{b,v})$ for each $v\in\affset$, where $c_{d,v}$ is a soft count of database items that 
have Hamming distance $d$ to the query and  are of the $v$-th affinity level.
Again, we use superscript $^{(i)}$ to indicate the dependency on $x_i$ as the query.

For a given $x_i$ and a given $v\in\affset$, we need to compute 
\begin{align}
{\alpha}^{(i)}_v = (\alpha_{0,v}^{(i)},\alpha_{1,v}^{(i)},\ldots,\alpha_{b,v}^{(i)})
= \left(\frac{\partial\mathcal{O}^{(i)}}{\partial c_{0,v}^{(i)}}, \frac{\partial\mathcal{O}^{(i)}}{\partial c_{1,v}^{(i)}}, \ldots, \frac{\partial\mathcal{O}^{(i)}}{\partial c_{b,v}^{(i)}} \right) \in \mathbb{R}^b.
\label{eq:alpha_vec}
\end{align}
Also, observe that the minibatch objective decomposes over the histogram bins, \ie, 
\begin{align}
    \mathcal{O}^{(i)} = \sum_{d=0}^b\mathcal{O}^{(i)}_d, 
\end{align}
for both AP \eqref{eq:ap_s} and DCG \eqref{eq:dcg_s}.
Since the $d$-th summand only depends on histogram bins  up to the $d$-th, we have that $\partial \mathcal{O}^{(i)}_d/\partial c_{l,v}^{(i)}=0$ when $d<l$.
As a result, \eqref{eq:alpha_vec} can be rewritten as
\begin{align}
{\alpha}^{(i)}_v 
& =  \left(
\frac{\partial\sum_{d\geq 0}\mathcal{O}^{(i)}_d}{\partial c_{0,v}^{(i)}}, 
\frac{\partial\sum_{d\geq 1}\mathcal{O}^{(i)}_d}{\partial c_{1,v}^{(i)}}, 
\ldots,
\frac{\partial\sum_{d\geq b}\mathcal{O}^{(i)}_d}{\partial c_{b,v}^{(i)}}
\right)\\
& = \left(
\frac{\partial\mathcal{O}^{(i)}_0}{\partial c_{0,v}^{(i)}}, 
\frac{\partial\mathcal{O}^{(i)}_1}{\partial c_{1,v}^{(i)}}, 
\ldots,
\frac{\partial\mathcal{O}^{(i)}_b}{\partial c_{b,v}^{(i)}} 
\right) + \left(
\sum_{d=1}^b\frac{\partial\mathcal{O}^{(i)}_d}{\partial c_{0,v}^{(i)}}, 
\sum_{d=2}^b\frac{\partial\mathcal{O}^{(i)}_d}{\partial c_{1,v}^{(i)}}, 
\ldots, 0
\right).
\label{eq:alpha_vec_s}
\end{align}

We next establish the fact that $\partial\mathcal{O}^{(i)}_d/\partial c_{l,v}^{(i)}$ is only a function of $d$ (independent of $l$) when $d\geq l$. First, the case of $d=l$ is trivial. 
When $d>l$, take $\mathrm{AP}_s$ \eqref{eq:ap_s} for example, omitting the superscript $^{(i)}$ for clarity:
\begin{align}
\frac{\partial \mathrm{AP}_s(\Rd)}{\partial c_{l}^+} & = 
\frac{\partial}{\partial c_{l}^+}\left(\frac{c_d^+}{N^+}\cdot \frac{C_{d-1}^++C_d^++1}{C_{d-1}+C_d+1}\right) \\
& = \frac{c_d^+}{N^+}\cdot \frac{2(C_{d-1}+C_d+1)-2(C_{d-1}^++C_d^++1)}{(C_{d-1}+C_d+1)^2} \\
& = \frac{c_d^+}{N^+}\cdot \frac{2(C_{d-1}^-+C_d^-)}{(C_{d-1}+C_d+1)^2}.
\end{align}
Clearly, this is only a function of $d$. 
The case for $\mathrm{DCG}_s$ can be similarly made. Now if we define two more shorthands,
\begin{align}
{\zeta}_{v}^{(i)} &  \overset{\Delta}{=} \left(
\frac{\partial\mathcal{O}^{(i)}_0}{\partial c_{0,v}^{(i)}}, 
\frac{\partial\mathcal{O}^{(i)}_1}{\partial c_{1,v}^{(i)}}, 
\ldots, 
\frac{\partial\mathcal{O}^{(i)}_b}{\partial c_{b,v}^{(i)}}
\right) \in \mathbb{R}^b,\\
{\theta}_{v}^{(i)} &  \overset{\Delta}{=} \left(
\,~~~~0~~~~,
\frac{\partial\mathcal{O}^{(i)}_1}{\partial c_{0,v}^{(i)}},
\ldots,
\frac{\partial\mathcal{O}^{(i)}_b}{\partial c_{b-1,v}^{(i)}}
\right) \in \mathbb{R}^b,
\end{align}
then  \eqref{eq:alpha_vec_s} can be further simplified as
\begin{align}
{\alpha}_{v}^{(i)} = {\zeta}_{v}^{(i)} +  U {\theta}_{v}^{(i)},
\label{eq:alpha_vec_s1}
\end{align}
where 
\begin{align}
U = \left[\begin{matrix}
0 & 1 & \cdots & 1 & 1 \\ 
0 & 0 &\cdots & 1 & 1 \\ 
\vdots & \vdots & \ddots & \vdots & \vdots \\ 
0 & 0 &\cdots & 0 & 1  \\ 
0 & 0 &\cdots & 0 & 0  \\ 
\end{matrix}\right] \in \mathbb{R}^{b\times b}
\end{align}
is an upper-triangular matrix of $1$'s with zero diagonal.

Finally, we consider the entire minibatch, where $i$ ranges from $1$ to $M$. Note that to construct the matrix $A_{d,v}$ in \eqref{eq:A}, we need to first form the vector $(\alpha_{d,v}^{(1)},\ldots,\alpha_{d,v}^{(M)})$.
Building upon \eqref{eq:alpha_vec_s1}, the following matrix equation is readily available:
\begin{align}
\left[{\alpha}_{v}^{(1)} \cdots {\alpha}_{v}^{(M)}\right]_{b\times M}
= \left[{\zeta}_{v}^{(1)} \cdots {\zeta}_{v}^{(M)}\right]_{b\times M}
+ U_{b\times b}
\left[{\theta}_{v}^{(1)} \cdots {\theta}_{v}^{(M)}\right]_{b\times M}. 
\end{align}
Then, the vector in question is simply the $d$-th row of this matrix.

\end{document}